\documentclass[11pt]{article}
\usepackage{coling2020}
\usepackage{times}
\usepackage{url}
\usepackage{latexsym}

\colingfinalcopy 

\makeatletter
\newcommand\footnoteref[1]{\protected@xdef\@thefnmark{\ref{#1}}\@footnotemark}
\makeatother
\makeatletter
\newcommand{\printfnsymbol}[1]{%
  \textsuperscript{\@fnsymbol{#1}}%
}
\makeatother

\title{Cross-lingual Inductive Transfer to Detect Offensive Language}
\author{
     Kartikey Pant \thanks{\hspace{2 mm} Both authors contributed equally to the work.} \and Tanvi Dadu \printfnsymbol{1} \\
    International Institute of Information Technology, Hyderabad \\
    Netaji Subhas Institute of Technology, New Delhi \\
    {\tt kartikey.pant@research.iiit.ac.in} \\
    {\tt tanvid.co.16@nsit.net.in} \\
}

\date{}
\usepackage{graphicx}
\begin{document}
\maketitle
\begin{abstract}
With the growing use of social media and its availability, many instances of the use of offensive language have been observed across multiple languages and domains. This phenomenon has given rise to the growing need to detect the offensive language used in social media cross-lingually. In OffensEval 2020, the organizers have released the \textit{multilingual Offensive Language Identification Dataset} (mOLID), which contains tweets in five different languages, to detect offensive language. In this work, we introduce a cross-lingual inductive approach to identify the offensive language in tweets using the contextual word embedding \textit{XLM-RoBERTa} (XLM-R). We show that our model performs competitively on all five languages, obtaining the fourth position in the English task with an F1-score of $0.919$ and eighth position in the Turkish task with an F1-score of $0.781$. Further experimentation proves that our model works competitively in a zero-shot learning environment, and is extensible to other languages.

\end{abstract}

\section{Introduction}
The prevalence of social media has made public commentary a critical aspect in shaping public opinion. Although freedom of speech is often advocated, offensive language in social media is unacceptable. Nevertheless, social media platforms and online communities are laden with offensive comments. This phenomenon results in the need for computationally identifying offense, aggression, and hate-speech in user-generated content in multiple languages.

This paper addresses the challenge put forward in the Multilingual Offensive Language Identification in Social Media shared task-organized at SemEval 2020 \cite{zampieri-etal-2020-semeval}. The theme of the problem is to identify the offensive language in tweets in Arabic, Danish, English, Greek, and Turkish. This shared task is further divided into three sub-tasks. The first task consists of the identification of offensive tweets in a multilingual setting, whereas the other two tasks consist of the categorization of offensive tweets and identification of targets in English.

Transfer learning is a methodology to utilize the knowledge acquired from one or more tasks to solve other related tasks. It is ubiquitous in the domain of Natural Language Processing. It can be classified into multiple types, like transductive transfer learning and inductive transfer learning. The former is used when the tasks are the same, but the corresponding domains are different, such as in the case of the cross-lingual learning for similar tasks. On the other hand, the latter is used when the tasks are different, but the domains are similar such as in the case of finetuning of pretrained contextual word embeddings.

Cross-lingual inductive learning has been used for many downstream tasks like multi-lingual variants of question answering, text classification, and text generation \cite{Artetxe2018MassivelyMS,Lample2019CrosslingualLM,Conneau2019UnsupervisedCR}. The rationale behind this is that the language with limited resources benefits from joint training over many languages. It also helps in performing zero-shot learning and handling of code-switched text, which otherwise are difficult to tackle. 

In this work, we propose the use of cross-lingual inductive learning to detect the offensive language in the given five languages. We use pretrained \textit{XLM-R} \cite{Conneau2019UnsupervisedCR} cross-lingual embeddings and train a single cross-lingual model for all five languages in the multilingual Offensive Language Identification Dataset. 

\section{Related Work}
\newcite{xu2012learning} proposed the task of identifying bullying in social media using NLP.  They presented benchmarking results for text classification among different NLP tasks using off-the-shelf solutions. Further, \newcite{Nobata2016AbusiveLD} proposed a machine learning-based method incorporating linguistic features, including n-gram, distributional semantic, and syntactic features, to detect the abusive language in user-generated online content. They also released a first of its kind corpus of user comments annotated for offensive language sampled from different domains.

\newcite{Zampieri2019PredictingTT} proposed the Shared Task OffenseEval 2019 for detecting offensive language in the tweets sampled from Twitter. They released the \textit{Offensive Language Identification Dataset} (OLID) consisting of $14,100$ tweets in English annotated for offensive content using a fine-grained three-layer annotation scheme. OffensEval 2020 \newcite{zampieri-etal-2020-semeval} uses the same dataset for the English task and similar data creation methodology for other languages. 

\newcite{Artetxe2018MassivelyMS} introduced an architecture for learning joint multilingual sentence representations, using a single BiLSTM encoder with a BPE vocabulary shared by all the languages. They use $93$ languages, spanning across $30$ different language families and $28$ different scripts for training the sentence representation embeddings. Experimental results on the XNLI dataset \cite{conneau2018xnli} for cross-lingual natural language inference and MLDoc dataset for cross-lingual document classification show that the cross-lingual transfer of learned features helps in improving the performance of the classification models in a cross-lingual setting.

\newcite{Chen2018UnsupervisedMW} proposed a fully unsupervised framework for learning multilingual word embeddings using only monolingual corpora. Unlike prior work in multilingual word embeddings, these embeddings exploit the relations between all the language pairs. By mapping all monolingual embeddings into a shared multilingual embedding space via a two-stage algorithm consisting of \textit{Multilingual Adversarial Training} and \textit{Multilingual Pseudo-Supervised Refinement}, the authors propose an effective method for learning the representation.

Recent works for the task of text classification exploit pretrained contextualized word representations rather than context-independent word representations. These pretrained contextualized word embeddings, like BERT \cite{Devlin2019} and RoBERTa \cite{2020roberta}, have outperformed many existing techniques on most NLP tasks with minimal task-specific architectural changes and training resources. Contextualized word embeddings have been used for detecting subjective bias \cite{subjective-detection}, and detecting hate speech in tweets \cite{Mozafari2019ABT}. 

To train multilingual variants of the contextualized word embeddings effectively, \newcite{Lample2019CrosslingualLM} proposed a cross-lingual language modeling methodology termed as \textit{XLM}. In their paper, they proposed two variants of their model: supervised and unsupervised multilingual language models. The supervised variant used parallel data with a new cross-lingual language model objective. \newcite{Conneau2019UnsupervisedCR} proposed \textit{XLM-R} extending \textit{XLM}. \textit{XLM-R} differs from \textit{XLM} in its ability to allow pretraining of the cross-lingual language models at scale using larger datasets and better-tuned hyperparameters. Their experiments on XNLI, MLQA, and NER show significant increases over the previous state-of-the-art. We further finetune \textit{XLM-R}, pretrained on $2.5$ TB Common Crawl corpus spanning $100$ languages. 

\begin{figure}[!ht]
        \center{\includegraphics[width=0.55\textwidth]
        {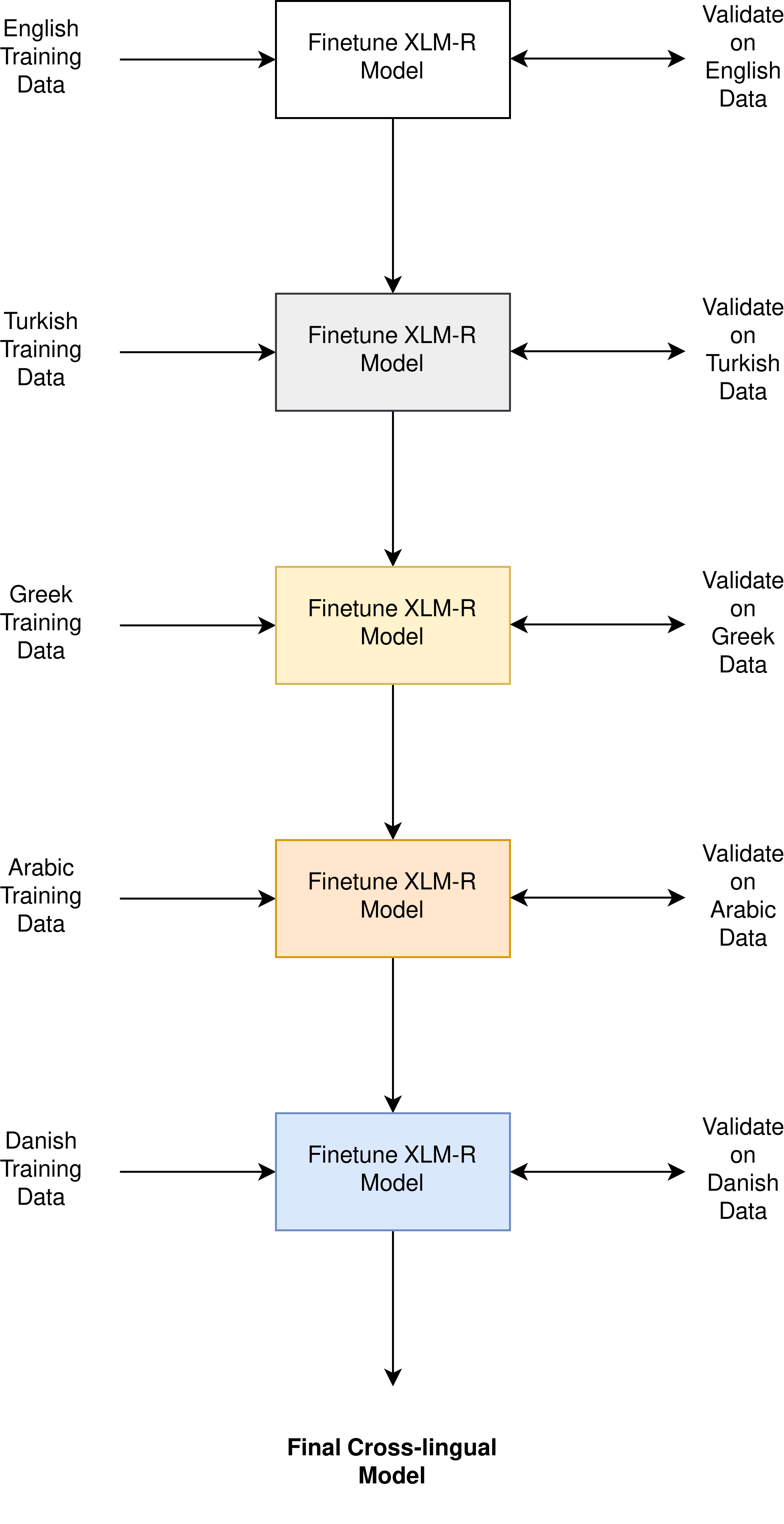}}
        \caption{\label{fig: architecture-diagram} Architecture of our cross-lingual model.}
\end{figure}

\section{Methodology}
In this section, we describe the preliminaries for the work: XLM-R, a cross-lingual contextual word representation, and cross-lingual inductive transfer. We then illustrate our sequential approach for fine-tuning the models.

\subsection{XLM-R}
\textit{XLM-R} is a transformer-based cross-lingual model pretrained using a multilingual masked language model objective on $2.5$ TB of CommonCrawl data in $100$ languages. \textit{XLM-R} obtains the state-of-the-art performance in cross-lingual classification, sequence labeling, and question answering. It obtains competitive results when compared with monolingual models on GLUE \cite{GluePaper} and XNLI tasks, showing that it is possible to have a single large model for all languages without sacrificing per-language performance.

\textit{XLM-R}, like other multilingual contextual word embeddings, suffers from the curse of multilinguality \cite{Conneau2019UnsupervisedCR}. While low-resource language performances can be improved by adding high resource languages using pre-training, the overall downstream task may suffer from capacity dilution \cite{Arivazhagan2019MassivelyMN}. This degradation happens because model capacity is constrained due to practical considerations including memory and speed during training. Moreover, for a fixed size model, the per-language capacity decreases with an increase in the number of languages. Despite these limitations, \textit{XLM-R} provides competitive results on downstream tasks when compared with mono-lingual models.

\subsection{Cross-lingual inductive transfer}
Multilingual contextual word embeddings retain a partial level of alignment, as has been observed in \cite{Cao2020Multilingual}. In \textit{XLM} and \textit{XLM-based} embeddings, multiple languages are specifically trained together with a sentence encoder leading to a higher degree of similarity in the alignment of corresponding words in different languages. Consequently, \textit{XLM} word embeddings have shown to perform competitively in SemEval’17 cross-lingual word similarity task \cite{Lample2019CrosslingualLM}. Thus, \textit{XLM} achieves a higher degree of alignment while learning cross-lingual representation. 

OffensEval 2020 provides a multilingual dataset for offensive language detection in five different languages, allowing one language’s learning to aid another. Therefore, cross-lingual inductive transfer learning using \textit{XLM-R} with relatively aligned embeddings allows for the inductive transfer of linguistic features among different languages.

Figure \ref{fig: architecture-diagram} illustrates our chain-like model for detecting offensive detection on all five languages trained in a sequential manner. While fine-tuning for every language, we validate the model only with the current language upon which it is being fine-tuned.

\section{Experiments and Results}
In this section, we present details on the multilingual OffenseEval dataset, experimental settings for the validation experiments, and specifications of the system runs.  

\subsection{Dataset}
In this subsection, we provide a comprehensive statistical analysis of the \textit{multilingual Offensive Language Identification Dataset} (mOLID). It comprises tweets in five different languages: English \cite{rosenthal2020}, Turkish \cite{coltekikin2020}, Greek \cite{pitenis2020}, Arabic \cite{mubarak2020arabic}, and Danish \cite{sigurbergsson2020offensive}. The tweets in English are annotated using a fine-grained three-layer annotation scheme, whereas the other four languages are annotated using a coarse-grained annotation scheme. For the English language, they use a three-level hierarchical annotation schema to distinguish between whether the language is offensive or not, the type of offensive language, and the target.

Statistical analysis of the dataset in Table \ref{tab:data-distribution} shows significant variation in the number of instances per language. Turkish has the maximum examples with $31,277$ examples, whereas Danish has $2,960$ examples, comprising $48.70\%$ and $4.60\%$ of total data, respectively. Further analysis shows that positive instances, signifying offensive languages, are significantly lesser than non-offensive sentences. Offensive sentences comprise only $23.21\%$ of total data. This percentage, however, varies from language to language, ranging from $12.97\%$ for Danish to $33.23\%$ for English.

\begin{table}[htbp!]
\centering
\resizebox{0.6\textwidth}{!}{%
\begin{tabular}{|l|r|r|r|}
\hline
\textbf{Language/Label} & \textbf{Positive} & \textbf{Negative} & \textbf{Total} \\ \hline
\textbf{English} & 4400 & 8840 & 13240 \\ \hline
\textbf{Turkish} & 6046 & 25231 & 31277 \\ \hline
\textbf{Greek} & 2486 & 6257 & 8743 \\ \hline
\textbf{Arabic} & 1589 & 6411 & 8000 \\ \hline
\textbf{Danish} & 384 & 2576 & 2960 \\ \hline
\textbf{Total} & 14905 & 49315 & 64220 \\ \hline
\end{tabular}%
}
\caption{Number of instances with associated label and language. }
\label{tab:data-distribution}
\end{table}
\subsection{Experimental Setting}
In this subsection, we outline the experimental setup for the task and present the results obtained on both the validation dataset and the blind test set. For experiments, we used \textit{XLM-R} having $16,550M$ parameters and $250K$ vocabulary size. For validation, we train and evaluate our model using $90-10$ train-validation data split for the five languages in the given order: English, Turkish, Greek, Arabic, and Danish. 

For models of all the languages, we finetune \textit{XLM-R} with a learning rate of $1*10^{-5}$ for $2$ epochs each with a maximum sequence length of $50$ and a batch size of $32$.  We evaluate all the models on the following metrics for the binary classification: \textit{F1}, \textit{Precision}, \textit{Recall}, \textit{Accuracy}.

\begin{table}[htpb!]
\centering
\resizebox{0.75\textwidth}{!}{%
\begin{tabular}{|l|r|r|r|r|}
\hline
\textbf{Language/Metric} & \textbf{F1 Score} & \textbf{Precision} & \textbf{Recall} & \textbf{Accuracy} \\ \hline
\textbf{English} & 0.709 & 0.728 & 0.690 & 0.808 \\ \hline
\textbf{Turkish} & 0.689 & 0.791 & 0.611 & 0.889 \\ \hline
\textbf{Greek} & 0.728 & 0.839 & 0.642 & 0.867 \\ \hline
\textbf{Arabic} & 0.729 & 0.762 & 0.698 & 0.907 \\ \hline
\textbf{Danish} & 0.676 & 0.686 & 0.667 & 0.915 \\ \hline
\end{tabular}%
}
\caption{Experimental Results for Validation dataset for our final model.}
\label{tab:experimental-results}
\end{table}

Table \ref{tab:experimental-results} shows the performance of our proposed model on the validation set for the five languages during the fine-tuning of the cross-lingual contextual embeddings. We observe that the performance metrics obtained vary significantly with the languages, with \textit{F1} varying from $0.676$ in Danish to $0.729$ in Arabic, and \textit{Accuracy} varying from $0.808$ in English to $0.915$ in Danish.

\begin{table}[htbp!]
\centering
\resizebox{0.3\textwidth}{!}{%
\begin{tabular}{|l|r|}
\hline
\textbf{Language/Metric} & \textbf{F1} \\ \hline
\textbf{English} & 0.919 \\ \hline
\textbf{Turkish} & 0.781 \\ \hline
\textbf{Greek} & 0.803 \\ \hline
\textbf{Arabic} & 0.848 \\ \hline
\textbf{Danish} & 0.759 \\ \hline
\end{tabular}%
}
\caption{Experimental Results for Test dataset for our final model.}
\label{tab:final-experimental-results}
\end{table}

Table \ref{tab:final-experimental-results} shows the performance of our proposed model on the test dataset held out by the organisers for all the five languages. Organisers used \textit{Macro-F1 score} for evaluating the model performance on test data. We observe that \textit{F1-score} varies from language to language, from $0.919$ from English to $0.759$ for Danish. 

\section{Discussion}

\textit{XLM-R} is an unsupervised cross-lingual representation pretrained using transformers on an enormous scale across $100$ languages. It can be fine-tuned for different downstream tasks in multiple languages, allowing one to take advantage of cross-lingual transfer learning. Like other transformer-based contextual word embeddings, \textit{XLM-R} is highly scalable, and the process of fine-tuning takes minimum training efforts and resources.  

Our proposed model, extending \textit{XLM-R}, was trained in five different languages for the detection of offensive language. Owing to the training of \textit{XLM-R} on hundred languages, our model allows a direct extension to other commonly-used languages. Our chain-like model exploits learnings from one language for detecting offensive language on another unrelated language. 

We further performed experiments for detecting offensive language in a language different from the ones that we have trained on without supervision. This form of learning, known as zero-shot learning, has been demonstrated in recent related works \cite{Artetxe2018MassivelyMS,Lample2019CrosslingualLM}. To evaluate the efficacy of our model to perform zero-shot learning, we detect offensive language text in German using our final cross-lingual model. We perform the experiments in the same experimental setting as the GermEval Shared Task on the Identification of Offensive Language’s Subtask 1 \cite{wiegand2018overview}, which entails coarse-grained binary classification of offensive language.

\begin{table}[htbp!]
\centering
\resizebox{0.75\textwidth}{!}{%
\begin{tabular}{|l|l|l|l|l|}
\hline
\textbf{Model/Metric} & \textbf{F1} & \textbf{Precision} & \textbf{Recall} & \textbf{Accuracy} \\ \hline
\textbf{\newcite{ParaschivCercel2019}} & 0.770 & 0.764 & 0.776 & 0.794 \\ \hline
\textbf{Our Model (Zero-Shot Learning)} & 0.700 & 0.795 & 0.682 & 0.781 \\ \hline
\end{tabular}%
}
\caption{Experiments performed for German Offensive Language Detection demonstrating zero-shot learning.}
\label{tab:discussion-experiment}
\end{table}
Our experiments, illustrated in Table \ref{tab:discussion-experiment} show results competitive to the best-performing model by  \newcite{ParaschivCercel2019}, which used supervised training data. Thus our model, despite not using supervised data in German, achieves competitive performance with a difference of $1.3\%$ in \textit{Accuracy}, further showing an increase in the metric of \textit{Precision} by a margin of $4.06\%$.

\section{Conclusion}
This work presents a cross-lingual inductive transfer learning approach to detect the offensive language in tweets across five languages: English, Turkish, Greek, Arabic, and Danish. Our proposed model performs competitively for all five languages. We further performed comprehensive experiments to show that our model performs well in a zero-shot setting, which can be useful in the case of low resource languages. Our proposed model can be easily extended to other languages with minimum training efforts and resources, performing competitively even with significantly imbalanced data. Future work may involve including multiple unrelated languages to enable a universal offensive language detection model and the application of the proposed approach in other cross-lingual tasks.

\bibliographystyle{coling}
\bibliography{coling2020,offenseval2020}

\end{document}